\documentclass{article}
\usepackage{aaai}
\nocopyright

\usepackage{amsmath}
\usepackage{amssymb}
\usepackage{mymacros}

\newcommand{\defp}{\specialchar{D}}
\newcommand{\pred}[1]{\ensuremath{Defined(#1)}}
\newcommand{\rules}[1]{\ensuremath{Rules(#1)}}
\newcommand{\state}{\specialchar{S}}
\newcommand{\CS}{\specialchar{CS}}

\begin{document}

\title{SLDNFA-system\thanks{http://www.cs.kuleuven.ac.be/${}_{\tilde{}}$dtai/kt/\mbox{systems-E}.shtml}}

\author{Bert Van Nuffelen \\
Department of Computer Science\\ Celestijnenlaan 200A\\ 3001 Heverlee, Belgium \\ bertv@cs.kuleuven.ac.be }
            
\maketitle

\begin{abstract}
\noindent 
The SLDNFA-system results from the LP+ project at the K.U.Leuven, which investigates logics and proof procedures for these logics for declarative knowledge representation.
Within this project inductive definition logic (ID-logic) is used as representation logic.
Different solvers are being developed for this logic and one of these is SLDNFA.
A prototype of the system is available and used for investigating how to solve efficiently problems represented in ID-logic.
\end{abstract}

\section{General Information}
The LP+ project at the K.U.Leuven aims at developing and investigating logics suitable for declarative knowledge representation. To be able to represent problem domains in a declarative way, the logic must be capable to express the knowledge of the expert in a natural and graceful way.
Therefore a suited logic has to deal with two mayor types of knowledge: definitional and assertional knowledge \cite{Denecker95a}.
This view is incorporated in ID-logic, a conservative extension of classical logic with a generalized notion of non monotone inductive definitions \cite{Denecker98c}.
In this setting, reasoning and problem solving in ID-logic is connected to model generation or satisfiability checking.
This is in general a undecidable problem, therefore the implemented system SLDNFA can handle only a restricted class of problems.
This class is still a large one: it involves scheduling, planning, simple theories like N-queens,\dots
The SLDNFA-system, which is one of the solvers developed in the project to reason on ID-logic theories, is an integration of abductive logic programming \cite{Kakas93a} and constraint logic programming \cite{JaffarMaher94}.
The current implementation is a prototype programmed as a meta program on top of Sicstus Prolog 3.7.1.


\section{Description of the System}

The SLDNFA-system consists of two layers: a general knowledge representation logic (ID-logic) and an actual problem solver (SLDNFA).
Before describing how these layers are connected to each other in the system, we first treat them separately.
\subsection{ID-logic} 
As mentioned, ID-logic is an extension of classical first order logic (FOL) with inductive definitions.
The logic builds upon the earliest ideas on the declarative semantics of logic programs with negation as failure.
The view of a logic program as a definition of its predicates is underlying both the least model semantics of van Emden and Kowalski \cite{vanEmden76} and Clark's completion semantics \cite{Clark78}.
This idea is further explored in \cite{Denecker98a}, where the authors argue that the well-founded semantics for logic programming implements a generalized principle of non-monotone induction.

Based on these ideas, an ID-logic theory $T$ is defined as consisting of a set of {\em definitions} and a set of classical logic sentences.
A definition is an expression that defines a subset of predicates in terms of the other predicates.
Formally, a definition \defp\ is a pair of a set \pred{\defp} of predicates and a set \rules{\defp} of rules that exhaustively enumerate the cases in which the predicates of \defp\ are true.
A rule is of the form: $ p(\ttt) \la F$ where $p \in \pred{\defp}$ and $F$ an arbitrary first order formula.
The predicates in \pred{\defp} are called {\em defined} by \defp, the others are called {\em open} in \defp.

The semantics of ID-logic integrates classical logic semantics and well-founded semantics. An interpretation $M$ is a model of a definition \defp\ iff it is total (i.e. 2-valued) and the unique well-founded model of \defp\ extending some interpretation $M_o$ of the functor and open predicate symbols of \defp. An interpretation $M$ is a model of an ID-logic theory $T$ iff it is a total model of its classical logic sentences and of its definitions.  Logical entailment is defined as usual: $T \models F$ iff $F$ is true in all models of $T$.

In the system, a definition is represented in a Prolog-like style using capitals for
variables, and ``\texttt{,}'', resp.  ``\texttt{;}'' for conjunction, resp. disjunction.
The \texttt{\$} is used as a delimiter for separating the quantifiers and the rest of the formulas. It doens't have a special meaning.
{\small
\begin{verbatimtab}[3]
uncle(X,Y)<- ( exists(Z)$ parent(Y,Z),brother(X,Z);
               exists(A)$ aunt(A,Y),married(A,X)  ).
\end{verbatimtab}
}
{\small
\begin{verbatimtab}[3]
aunt(X,Y) <- ( exists(Z)$ parent(Y,Z),sister(X,Z);
               exists(A)$ uncle(A,Y),married(A,X)).
\end{verbatimtab}
}
\noindent The above  definition defines the two predicates \textsl{uncle} and \textsl{aunt} simultaneously. The other are open predicates.

FOL axioms are represented in the system in the same style but are prefixed by the key-word {\tt fol}. They are a straightforward representation of FOL.
{\small
\begin{verbatimtab}[3]
fol forall(X,Y)$ 
		uncle(X,Y), age(X,AgeX), ageY(Y,AgeY)
      => AgeX > AgeY.
fol aunt(mary,bob).
\end{verbatimtab}
}

ID-logic generalizes not only classical logic but also abductive logic
programming \cite{Kakas93a} and open logic programming
\cite{Denecker95a}. An abductive logic framework, consisting of a set
of abducible predicates, a set of rules and a set of FOL {\em
  constraints} can be embedded in ID-logic as the theory consisting of
the FOL constraints and one definition defining all non-abducible
predicates. Formally, ID-logic extends ALP by allowing multiple
definitions and generalized syntax. However, it can be shown that it
is always possible to transform a set of definitions into one single
definition. 

{\bf A special note} The definitions we consider are well-founded \cite{Denecker00}. The
models of a well-founded definition coincides with the well-founded semantics of the models of the
completion of the definition.  Below, completed definitions of predicates will be denoted: $\forall(p(\overline{X}) \lra B_p[\overline{X}])$.

\subsection{SLDNFA}
Given an ID-logic theory $T$ containing one definition \defp , an abductive problem for a given query $F$\footnote{Note that in contrast to Logic Programming conventions a query is stated positively.} consists of computing a definition $\Delta$ of ground atoms for
the open predicates of $T$ and an answer substitution $\theta$ such
that $\defp+\Delta$ is consistent and entails all FOL axioms in $T$ and
$\forall(\theta(F))$. An abductive procedure computes tables for the open predicates that can be extended in a unique way to a well-founded model of the definition, and a model of the FOL axioms and the query.

SLDNFA \cite{Denecker92d,Denecker98z} is an abductive procedure for normal logic programs.  The procedure sketched below is an extension of it to deal with FOL axioms and generalized rules and queries.  As in the case of \cite{Fung97} and \cite{Mantsivoda96}, the procedure is proposed as a set of rewrite rules. 

A derivation for a query $F$ is a rewriting process of {\em states $\state$}, i.e. tuples $(\Theta,\Delta,\CS)$ of a set $\Theta$ of FOL formulas and denials, a set $\Delta$ of abduced atoms and a constraint store .
A denial is a formula of the form $\forall \overline{X}. \la F[{\overline{X},\overline{Y}}]$, where $\la$ denotes negation.
Denials are the only formulas that may contain universal quantifiers.
In general a theory can be transformed into denials applying the following rewriting step $\forall\ X.F \ra \neg \exists\ X. \neg F$.
Open variables in FOL formulas and denials represent objects of yet unknown identity.

The initial state is the pair $(\Theta,\emptyset,\emptyset)$ where $\Theta$ consists of $F$ and the set of FOL axioms in $T$. The rewriting process proceeds by selecting a formula $G$ from $\Theta$ and computing a new state according to rewrite rules as explained below. 
If $fail$ or an inconsistent constraint store is derived during the computation, the computation backtracks.  
The computation ends in three possible ways:
\begin{itemize}
\item with \textsl{failure}, if no solution is derived;
\item with a \textsl{successful} derivation if a state $\state$ is derived which contains a consistent constraint store \CS\ and 
  $\Theta$ only consists of negative goals of the following form:
\begin{itemize}
\item
$\forall \XX \la Y=t \land Q$ where $Y$ is a free variable and $t$ any term not in $\XX$
\item $\forall \XX \la a(\ttt) \land \neg (\ttt=\sss_1 \lor ...  \lor \ttt=\sss_n) \land Q$ where $\{a(\sss_1),..,a(\sss_n)\}$ is the
  set of abduced $a$-predicates in $\Delta$.
\end{itemize}
An answer consists of the substitution of the free variables of the
initial query and of the set of abduced atoms $\Delta$.
\item with a \textsl{floundering} error condition when universally quantified
  variables appear in a selected negative literal.
\end{itemize}

During the derivation a formula $G$ is selected from a the set $\Theta_i$ in a state $(\Theta_i,\Delta,\CS)$.
According to the rewrite rules we obtain a new state $(\Theta',\Delta',\CS')$. 
In table \ref{SLDNFA-rules} the basic rules are displayed. 
If $G$ is a formula containing a CLP-expression as first literal then special rules, shown below, are applied.
In the rules we use the notation $A + B$ as a shorthand for $\{A\} \cup B$.
$sf(\theta)$ denotes the solved form of the substitution $\theta$.
The literal $a(\XX)$ denotes an open predicate, while $p(\XX)$ a defined one.
For the sake of clarity we write only the component of the state that changes, in terms of the components of the old state. 
Let $\Theta_i = \Theta \cup \{G\}$.

\begin{table*}[t]
\begin{center}
\begin{tabular}{ll}
\begin{minipage}[t]{5cm}
\[\begin{array}{lll}
{\bf true }& \rewriteto & \Theta\\
{\bf false }& \rewriteto & fail\\

{\bf p(\ttt)  }& \rewriteto  & B_p[\ttt] + \Theta \\

{\bf a(\ttt) }& \rewriteto & a(\ttt) + \Delta\\

{\bf F\land G   }&  \rewriteto &  F + G + \Theta  \\

{\bf F\lor G  }&  \rewriteto  & \left\{ 
\begin{array}{c}
F + \Theta  \\ 
    \mbox{  or  } \\
G  + \Theta  \\ 
\end{array} \right .\\

{\bf \exists X. F[X]  }& \rewriteto &  F[X] + \Theta  \\

{\bf \neg F   }&  \rewriteto  &  \leftarrow\!F + \Theta   \\ 
\end{array}\]
\end{minipage}
 &
\begin{minipage}[t]{7cm}
\[ \begin{array}{lll}
{\bf \leftarrow true  }& \rewriteto & fail\\

{\bf \leftarrow false }& \rewriteto & \Theta\\

{\bf \forall \XX \la p(\ttt) \land Q   }& \rewriteto & 
        \forall \XX \la B_p[\ttt] \land Q\ \  + \Theta  \\

{\bf 
\forall \XX \la (F\lor G) \land Q  }& \rewriteto &
\begin{cases}        \forall \XX \la F \land Q\ \  + \\
        \forall \XX \la G \land Q\ \   + \Theta 
\end{cases}\\

{\bf \forall \XX \la (\exists \YY F \land Q) }& \rewriteto &
                 \forall \XX,\YY \la F \land Q\ \ + \Theta \\

{\bf \forall X,\YY  \la X=t \land Q[X]  }& \rewriteto &
        \forall \YY \la Q[t]\ \  + \Theta \\

\end{array}\]
\end{minipage}
\end{tabular}
\end{center}

\[\begin{array}{lll}

{\bf \sss=\ttt  }& \rewriteto &
\begin{cases}
\theta((\Theta,\Delta)) &  \text{ if $\theta$ is m.g.u. of $\sss=\ttt$,} \\
 fail& \text{   if  $\sss=\ttt$ has no m.g.u.}
\end{cases} \\
\\
{\bf \forall \XX \la \sss=\ttt \land Q  }& \rewriteto &
\begin{cases} 
  \forall \XX \la sf(\theta)  \land Q\ \   + \Theta &  \text{ if $\theta$ is m.g.u. of    $\sss=\ttt$  ,}\\
  \Theta & \text{ if $\sss=\ttt$ has no m.g.u. }
\end{cases}\\
\\
{\bf \forall \XX \la a(\ttt) \land Q   }&  \rewriteto & 
\begin{cases}
\forall \XX \la \sss=\ttt \land Q + \\
\forall \XX \la a(\ttt) \land \neg \sss=\ttt \land Q\ \  + \Theta \\
\mbox{ for some $a(\sss) \in \Delta$} 
\end{cases}\\
\\
{\bf \forall \XX \la \neg F \land Q   }&  \rewriteto & \left \{
\begin{array}{l}
\mbox{if $Free(F) \cap \XX = \emptyset$ then }\\
\mbox{then} \left \{ \begin{array}{c}
                      F\ \  +  \Theta \\
                      \mbox{ or }\\
                      \leftarrow\! F\ \  +  \forall \XX \la Q\ \  + \Theta
                     \end{array} \right . \\
\mbox{otherwise a floundering error condition occurs}
\end{array}
\right .
\end{array}
\]
\caption{The rules of the SLDNFA-procedure}\label{SLDNFA-rules}
\end{table*}
If $G$ contains a CLP-expression ${\bf C}$ then the following rules needs to be applied instead of the rules in table \ref{SLDNFA-rules}. 
\[
\begin{array}{l}
{\bf C[\YY]}  \hspace{3.2cm}\rewriteto C[\YY] + \CS \\
{\bf \forall\ \overline{X} \la\!C[\YY] \land B}\\
\rewriteto 
\left\lbrace\begin{array}{l}
(\Theta, \Delta, \neg C[\YY] + \CS) \\
\hspace{1.5cm}\text{or}\\
(\forall\ \overline{X} \la\!B  + \Theta, \Delta, C[\YY] + \CS)
\end{array}\right.\text{if $\YY\cap \XX = \emptyset$}\\
\\
{\bf \forall\ \overline{X} \leftarrow C[\overline{X},\overline{Y}] \land B[\overline{X}]} \hspace{.4cm}\rewriteto floundering 
\end{array}
\]
The addition of a CLP-expression to the constraint store may lead to an inconsistent constraint store. 
If this is the case another branch of the rule will be applied. If none succeeds the rule derives $fail$. 
For readability this is not stated in the rules.

The last rule above states that a mixing of free variables and universal quantified variables leads to floundering.
However in two specific cases this isn't the case: first when the selected CLP-literal contains only universal quantified variables, and second when the CLP-literal is an equality.
\[
\begin{array}{l}
{\bf \forall\ \overline{X} \la\!C[\XX] \land B[\overline{X}]} \\%
\rewriteto \left \{
\begin{array}{l}
\mbox{Let $Sol$ be the set of solutions of $C[\XX]$}\\
\left \{\begin{array}{ll}
\{ \forall \XX. \la\!B[\sss]\ \| \ \sss \in Sol \}  \cup \Theta
& \text{if $Sol$ is finite}\\
floundering
  & \text{if $Sol$ is infinite }
\end{array}\right.\end{array}\right.\\
\\

{\bf \forall\ X,\overline{Y} \la X = Z \land B[X,\overline{Y}] }\\
\rewriteto \forall\ \overline{Y} \la B[Z,\overline{Y}] 
\end{array}
\]
In the first rule a universal quantified CLP-literal is selected in some denial.
If the literal has a finite number of solutions (e.g. $\forall\ X \la X\ in\ 1..10\land Q[X]$), these solutions are collected and for each solution the resulting denial is added.

\subsection{The implemented system}
The implemented system consists of two parts: a preprocessor and the actual reasoner.
The preprocessor will take a typed ID-logic specification and check it for syntax and type errors. 
Then it will transform the specification towards a form which is handled by the abductive solver.
Types \cite{DeMot99} were added to ID-logic theories for a number of reasons: to reduce the number of simple and stupid errors made by experts, to disambiguate expressions (e.g. between real or natural numbers) and  to generate more optimized transformed expressions. 
The current type system can handle simple many sorted types and is able to infer types for predicates for which a type declaration is lacking.

Given a transformed specification the abductive solver will reason on it and generate an answer according to a query.
During the computation the solver will try to avoid as much as possible backtracking either by postponing goals or by pushing as much as possible of the backtrack points into the CLP-constraint store which is constructed during the derivation. 
The latter is done because a CLP solver backtracks much faster than SLDNFA can.

\section{Applying the System}


\subsection{Methodology}

                                        
The task to define a problem for the system is actual the (declarative) knowledge representation task.
The user first defines the ontology of its problem domain: the relevant types of objects, the relevant relations and functions between them.
Then he chooses a logical alphabet to name them.
Using this alphabet, he will express his knowledge by a set of logical sentences that are true in the problem domain. 
To illustrate this proces, an example is elaborated in the appendix.
The above sketched methology is a more or less theoretical one, in practice adding and retracting concepts and axioms are often interleaved.

\subsection{Specifics}
The system is developed from the view that logic is a very flexible and accurate formalism to represent knowledge.
We believe that if an expert follows the above sketched methodology the chosen representation often leads to  model generation or satisfiability checking. 
As this is related to abductive reasoning a solver which performs abduction in a certain way is needed.


\subsection{Users and Usability} 
The most important requirement for a potential user is that he's able to formalize his knowledge about his problem domain in logic.
Sometimes this is straightforward, but in other cases it can be very hard.
To facilitate the user, shorthand declarations for some typical knowledge patterns, like a function from which the domain is known and the range values are unknown, are introduced.
Also the logic is enhanced with some higher order declarations involving aggregates. 
This allows the user to specify in a convenient way aggregation knowledge.
Except from the logical representation the user has to declare some extra information needed for the system: type information, open predicate declaration and etc \dots

The practical use of the system is somehow restricted.
The current implementation is a meta program which slows down the computation and don't allow to do huge experiments.
Moreover the solver can only deal with a restricted class of problem specifications.
In some cases the solver will go in an infinite loop (e.g. transitive closure) or need an endless time to find a solution (e.g. planning problems).
We try to broaden this class by introducing new resolution techniques (e.g. tabling) or integrating it with other systems.
Even with this restriction, the solver can tackle a wide range of problems, including scheduling, planning, satisfiability checking, \dots


\section{Evaluating the System}


\subsection{Benchmarks}
The SLDNFA system is constructed to investigate how to build efficient reasoners on declarative knowledge representations.
Therefore next to computing efficiency, the benchmarks should include maintainability, adaptability and extendibility of the problem specification.  
Also the time needed to represent the problem is an important factor.
As well known, these factors are less easy to quantify as performance differences.
In general, one should evaluate this system keeping in mind its expressivity power and its ability to solve certain classes of problems in a reasonable time.


\subsection{Comparison} 
Because the SLDNFA-system is a general purpose system, it probably  will always be outperformed by special purpose systems in a specific domain.
But mostly these systems aren't able to deal with a broad class of problems or give the same flexibility, maintainability and expressivity.
However there exists comparative systems (solvers): e.g. sModels \cite{sModels} and ACLP \cite{ACLP}. 
These systems can be used as alternative solver instead of the SLDNFA procedure if the appropriate transformation is applied on ID-logic theories. 
Compared to them we offer in some cases a better performance, in other a more flexible representation. An extended comparison can be found in \cite{Pelov00}.

An interesting class of problems to compare with is the one which traditionally is developed in CLP.
This is because the current SLDNFA-system reduces a specification of such problems into a CLP-constraint store.
The results for some examples like the N-queens problem, graph coloring and job-shop scheduling show that SLDNFA has a polynomial overhead compared to pure CLP-representations.
It constructs less specialized and less optimized constraint stores.
But clearly this is a consequence of the high level knowledge representation.

\subsection{Problem Size} 
At the moment, the largest experiment done is the scheduling of maintenances of units a Belgian electricy provider.
The theoretical search space of this problem consists of $52^{48}$ states. 
The SLDNFA-system was able to find the same optimal solution of the problem as a pure CLP-program in about 20 minutes.
The (optimize) CLP-program only used 3 minutes.
More details can found in \cite{VanNuffelen00}.

In the domain of planning, we tried to solve some simple queries.
Some could be solved in a very fast way, in others situations the solver got lost in a huge backtracking process.
But compared to earlier versions of the system, the performance boost-up is significant.
For example reversing a 6 high tower in the blocks world was impossible with the previous versions, now the current system is able to reverse a 65 high tower within one hour.
Some more examples can be found in \cite{Pelov00,VanNuffelen99}.

\section{Acknowledgements}
Bert Van Nuffelen is supported by the GOA LP+ project at the K.U.Leuven. The SLDNFA-system is implemented by the ideas of Marc Denecker, Danny De Schreye and many others. 
In particular the preprocessor is the work of Emmanuel De Mot.

\appendix
\section{The N-queens puzzle}
In the queens problem, N queens have to be placed on a N by N board, so that no queen attacks another.
Because we intuitively know that a solution to the puzzle contains no two queens on the same column, it is natural to associate a column to a queen. 
As consequence, the row position of a queen can be simple represented by a predicate {\tt has\_position(i,j)} which expresses that the queen {\tt i} at the {\tt i}'th column is   on row {\tt j}.
Together with  the predicate {\tt dim(N)}, which represents the size of the puzzle, we have defined the ontology of the problem.
Typing these concepts is easy: in both predicates the arguments range over the integers (\texttt{int}).
The predicates them self have the type \texttt{pred}.
Below a renaming of the integer type is done to show how type information can enhance the readability of the specification. 
{\small
\begin{verbatim}
type_instance(pos,int).

has_position(pos,pos)::pred.
dim(int)::pred.
\end{verbatim}
}

Before representing the axioms how the queens stand in relation to each other, we first define an auxiliary concept: namely the domain of a row or a column. 
The second definition defines that the size of the considered puzzle is 8.
{\small
\begin{verbatim}
dom(X) <- dim(N), X in 1..N.
dim(8) <- true.
\end{verbatim}
}
Note that {\tt has\_position(i,j)} has no definition therefore it should be declared an open predicate by the declaration
{\small
\begin{verbatim}
abducible(has_position(_,_)).
\end{verbatim}
}
Based on the above defined concepts we are able to represent the puzzle by the following sentences.
\begin{itemize}
\item Each queen has a position on the board.
{\small
\begin{verbatim}
fol forall(Q,P) $ 
    dom(Q) 
    => (exists(P) $  dom(P), has_position(Q,P)).
\end{verbatim}
}
\item Each queen is positioned on at most one row
{\small
\begin{verbatim}
fol forall(Q,P1,P2) $ 
    has_position(Q,P1), has_position(Q,P2) 
    => P1 = P2.
\end{verbatim}
}
\item Two different queens are on different rows
{\small
\begin{verbatim}
fol forall(Q1,Q2,P1,P2) $ 
    has_position(Q1,P1), has_position(Q2,P2), 
    Q1 \= Q2 
    => P1 \= P2.
\end{verbatim}
}
\item Two different queens are on different diagonals
{\small
\begin{verbatim}
fol forall(Q1,Q2,P1,P2) $ 
    has_position(Q1,P1), has_position(Q2,P2), 
    Q1 \= Q2 
   => Q1 + P1 \= Q2 + P2,  Q1 - P1 \= Q2 - P2.
\end{verbatim}
}
\end{itemize}

Two remarks apply to this representation: a computational and a representational one. 
First you can see that the last two axioms overlap in the condition of the implication, so its better to combine these; also these axioms are symmetrical and clearly these axioms can be represented in an asymmetrical form.
Thus we can replace both  axioms by the following axiom:
{\small
\begin{verbatim}
fol forall(Q1,Q2,P1,P2) $ 
    has_position(Q1,P1), has_position(Q2,P2), 
    Q1 < Q2
    => P1 \= P2, Q1 + P1 \= Q2 + P2, 
       Q1 - P1 \= Q2 - P2.
\end{verbatim}
}
This meta reasoning on the representation is at the moment left to the expert. 
However this effort results a huge difference in computation time.

The second remark concerns a property of the \texttt{has\_position} relation.
Observe that {\tt has\_position} is a bijection in $\{1,..,N\}$.
As previous mentioned, for some knowledge patterns we introduced special notations.
Applied on the N-queens representation, we obtain:
{\small
\begin{verbatim}
type_instance(pos,int).

has_position(pos,pos)::pred.
dim(int)::pred.

dim(8) <- true.
dom(X) <- dim(N), X in 1..N.

ob has_position :: dom(_) -> dom(_).

fol forall(Q1,Q2,P1,P2) $ 
   position(Q1,P1), Q1 < Q2, position(Q2,P2) 
   => Q1 + P1 \= Q2 + P2,  Q1 - P1 \= Q2 - P2.
\end{verbatim}
}
Although not essential, the introduction of such high level declarations enhances the readability of the represention.
Further more it allows an efficient special purpose treatment by the solver.

A solution of the puzzle is a model which satisfies the above theory.
This can be obtained by quering the system with the query {\tt true}.
The model is represented by the table of \texttt{has\_position} atoms.

\bibliography{/home/bertv/tekst/Papers/bertlib,/home/bertv/tekst/Papers/marclib}
\bibliographystyle{/home/bertv/tekst/LaTeX/aaai}

\end{document}